\PassOptionsToPackage{table,dvipsnames}{xcolor}

\documentclass[10pt,twocolumn,letterpaper]{article}

\usepackage{cvpr}              
\usepackage{multirow}
\usepackage{graphicx}
\usepackage{caption}
\usepackage{array}
\usepackage{adjustbox}
\usepackage{booktabs} 
\renewcommand{\arraystretch}{1.2} 
\usepackage{threeparttable}
\usepackage{blindtext}
\usepackage[utf8]{inputenc}
\usepackage[T1]{fontenc}
\usepackage{amsmath}
\usepackage{amsfonts}
\usepackage{amssymb}
\usepackage{subcaption}
\usepackage{graphicx}
\usepackage{booktabs}
\usepackage{soul}
\usepackage{multirow}
\usepackage{soul}
\usepackage{bbm}
\usepackage{diagbox}
\usepackage{bbm}
\usepackage[utf8]{inputenc}
\usepackage{amsmath, amssymb}
\usepackage{algorithm}
\usepackage[noend]{algpseudocode}
\usepackage{booktabs}
\usepackage{graphicx}

\DeclareCaptionLabelFormat{nocaption}{}

\usepackage[dvipsnames]{xcolor}

\definecolor{cvprblue}{rgb}{0.21,0.49,0.74}
\usepackage[backref,breaklinks,colorlinks,citecolor=cvprblue]{hyperref}
\usepackage[capitalize]{cleveref}

\crefname{section}{Sec.}{Secs.}
\Crefname{section}{Section}{Sections}
\Crefname{table}{Table}{Tables}
\crefname{table}{Tab.}{Tabs.}
\def\paperID{12456} 
\def\confName{CVPR}
\def\confYear{2025}

\title{Plug-and-Play Interpretable Responsible Text-to-Image Generation\\ via Dual-Space Multi-facet Concept Control}
\vspace{-3mm}

\author{
Basim Azam \quad Naveed Akhtar\\
School of Computing and Information Systems, The University of Melbourne, Australia\\
{\tt\small \{basim.azam, naveed.akhtar1\}@unimelb.edu.au}
}

\vspace{-3mm}

\begin{document}


\twocolumn[{%
    \renewcommand\twocolumn[1][]{#1}%
    \setlength{\tabcolsep}{0.0mm} 
    \newcommand{\sz}{0.125}  
    \maketitle
    \begin{center}
        \newcommand{\teaserwidth}{\textwidth}
    \vspace{-0.5em}
        \includegraphics[width=\linewidth]{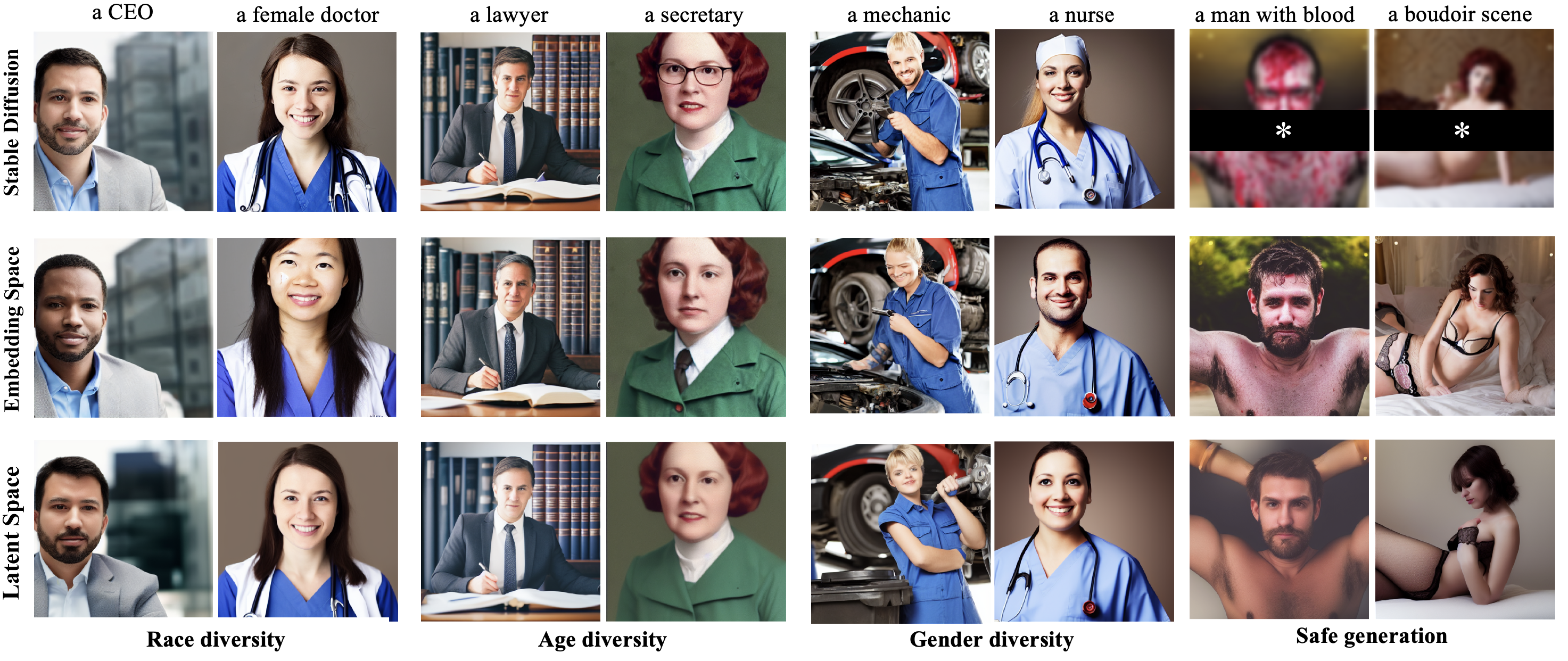}
      \vspace{-5mm}
        \captionof{figure}{Our unique plug-and-play interpretable approach simultaneously controls a range of concepts for responsible and fair image generation with text-to-image pipelines. Our method enables control over both text embedding space and latent diffusion space. Shown examples compare unfair and unsafe generation for the given prompts by the Stable Diffusion (top), along with their responsible counterparts resulting from our approach influencing the text encoder (middle) and the diffusion model (bottom). We  control individual concepts for diverse/safe generation in these examples while modeling them in continuous composite responsible semantic spaces.}
    \label{fig:first}
    \end{center}
}]
\maketitle

\begin{abstract}
\vspace{-3mm}
Ethical issues around text-to-image (T2I) models demand a comprehensive control over the generative content. Existing techniques addressing these issues for responsible T2I models aim for the  generated content to be fair and safe (non-violent/explicit). However, these methods  remain bounded to handling the facets of responsibility concepts individually, while also lacking in interpretability. Moreover,  they often require alteration to the original model, which compromises the model performance. 
In this work, we propose a unique technique to enable  responsible T2I generation by simultaneously accounting for an  extensive range of concepts for fair and safe content generation in a scalable manner. The key idea is to distill the target T2I pipeline with an external plug-and-play mechanism that learns an interpretable composite responsible space for the desired concepts, conditioned on the target T2I pipeline. We use knowledge distillation and concept whitening to enable this. At inference, the learned space is utilized to modulate the generative content. A typical T2I pipeline presents two plug-in points for our approach, namely; the text embedding space and the diffusion model latent space. We develop modules for both points and show the effectiveness of our approach with a range of strong results. Our code can be accessed at \textcolor{magenta}{https://basim-azam.github.io/responsiblediffusion/}      
\end{abstract}

\vspace{-5mm}
\section{Introduction}
\label{sec:intro}
\vspace{-1mm}
Rapid advances in text-to-image (T2I) generative pipeline are revolutionizing numerous vision applications, offering unprecedented convenience  for synthesizing high quality visual content using textual descriptions \cite{rombach2022high,ramesh2022hierarchical, peebles2023scalable, karras2022elucidating,ho2020denoising}. 
Unfortunately, this convenience coupled with the generated content realism offered by  publicly accessible T2I models, e.g., Stable Diffusion \cite{rombach2022high}, DALL-E \cite{ramesh2022hierarchical},  Imagen \cite{saharia2022photorealistic}, also has undesired ethical implications.
Confronting the core values of fair and safe (non-violent/explicit) development and deployment of Artificial Intelligence, these models  lack in producing  generative content responsibly~\cite{gu2023towards,zhang2025generate}. 

The reasons of irresponsible content generation by  contemporary T2I models lie deep in their training process, which must use large volumes of (mostly uncurated) data.  Issues like inappropriate content and harmful stereotyping lurking in the data are passed on to the generative models through   training~\cite{ba2023surrogateprompt,tsai2023ring}, which cause problems after their deployment.   
Hence, it is critical to develop methodologies to explicitly control  generative models from producing irresponsible content. 
Addressing that, several techniques have emerged recently, encompassing approaches like  input prompt filtering \cite{yoon2024safree,ba2023surrogateprompt}, post-hoc content moderation \cite{yang2024guardt2i, wang2024moderator, kumari2023ablating}, machine unlearning \cite{wu2024unlearning, park2024direct, huang2023receler, zhang2024forget}, and model editing  \cite{gandikota2024unified, xiong2024editing}. Though effective, these approaches face limitations like requiring human-intervention, ad-hoc methodology, and selective dealing of different facets of the concepts defining `responsible generation'.     

Another contributing factor to the ethical concerns related to T2I models is our current lack of  wholesome understanding of the latent space of the diffusion models. 
Researchers have already started exploring  interpretability \cite{haas2024discovering, zhang2024steerdiff} and manipulation of diffusion model latent spaces to mitigate unintended T2I model behavior \cite{gandikota2024unified, li2024self}. 
Controlling the latent space holds promise when it is possible to disentangle semantic concepts in it. One recent inspiring approach in this direction \cite{li2024self}  identifies vectors in the latent diffusion space in the directions of individual semantic concepts, which can be used to restrict model outputs. However, it is arguable that such a discretized treatment of the concepts within the diffusion latent space is restrictive. It is also intrinsically limited to dealing with diverse  facets of the concepts individually. 
Overall, it still remains a widely open challenge for the research community to comprehensively embed the multi-faceted concepts related to fair and safe image generation in T2I pipelines. 


In this work, we address this by presenting a unique scalable approach to extensively  incorporate fair and safe generative abilities in a T2I pipeline in a plug-and-play manner.
Our key insight is that a T2I model can be distilled for a range of user-desired responsibility concepts,  conditioned on the original model. We perform this distillation by inducing a student model that treats the concepts related to fairness and safety in a continuous space. Within this space, we further leverage concept whitening~\cite{Chen2020cynthia} to disentangle the representations of concepts for a better ultimate control on the generative image modulation. The underlying framework of our technique is generic, in that it is applicable to any representation space encoding semantic concepts. Hence, we explore its application to both text encoder embedding space and diffusion model latent space in the T2I pipeline. Methods for both variants address their unique challenges, but prove equally effective - see Fig.~\ref{fig:first}. The key contributions of this work are summarized below.

\begin{itemize}
    \item We propose a unique plug-and-play method for responsible image generation with T2I pipeline that enables comprehensive incorporation of fairness and safety concepts while modeling them under a distilled continuous space conditioned on the generative model.
    \item We tailor our method as a text embedding plug-in, and  diffusion latent plug-in while also leveraging concept whitening to ensure precise interpretable control over the content. Both plug-ins are employed in our approach.
    \item With extensive experiments, we not only demonstrate state-of-the-art or comparable performance in fair and safe image generation, but also show other unique interesting properties of our approach. 
\end{itemize}

\section{Related Works}
\label{sec:related}

Text-to-image (T2I) generation has significantly expanded the capabilities of generative AI, enabling high-quality outputs using textual descriptions \cite{rombach2022high, ramesh2022hierarchical, peebles2023scalable, karras2022elucidating, ho2020denoising}. While T2I methods like Stable Diffusion \cite{rombach2022high}, DALL-E \cite{ramesh2022hierarchical}, GLIDE \cite{nichol2021glide}, and Imagen \cite{saharia2022photorealistic} have achieved impressive realism and adaptation in numerous applications, they also face several challenges. One of them is the concern about their irresponsible and unethical content generation \cite{gu2023towards, ba2023surrogateprompt}. 

Contemporary T2I models are known to generate inappropriate images, including nude and violent content, and they are also susceptible to social and cultural biases persistently found in the public domain datasets used to train these models~\cite{cho2023dall, luccioni2024stable, struppek2022biased}. The increased popularity of T2I models has driven research into understanding and mitigating biases in their outputs, particularly focusing on safety filters \cite{rando2022}, semantic space organization \cite{brack2022stable}, and tendencies toward stereotypical portrayals \cite{bianchi2022easily, cho2023dall, naik2023social}. Studies show that these models, often guided by biased systems, e.g., CLIP \cite{radford2021learning, agarwal2021evaluating}, may memorize sensitive data \cite{carlini2022, somepalli2022diffusion}, which, combined with unfiltered web-based datasets containing harmful content \cite{birhane2021multimodal, siddiqui2022metadata, berg2022prompt, bansal2022well}, propagates unintended biases.  

Research to address these concerns can be divided into three main directions; namely, preemptive content filtering, model-level adjustments, and post-hoc moderation. The preemptive filtering approaches involve controlling input prompts to reduce biases before they influence image generation. For example, prompt-based filters \cite{yoon2024safree, ba2023surrogateprompt} screen out language associated with inappropriate content \cite{karras2022elucidating, peebles2023scalable}. 
In the model-level adjustment, unlearning techniques \cite{wu2024unlearning, park2024direct, huang2023receler, zhang2024forget} aim to remove sensitive concepts directly from the model’s learned representations. Similarly, model editing methods  target specific parameters of the models to limit the generation of harmful content without extensive re-training \cite{orgad2023editing, gandikota2024unified, xiong2024editing}. Although useful, these methods can negatively impact the original model  performance. They can also become computationally intensive as the harmful concepts increase~\cite{zhang2024forget}. Post-hoc moderation methods \cite{yang2024guardt2i, wang2024moderator, kumari2023ablating} involve  both automated and manual processes like GuardT2I \cite{yang2024guardt2i} and context-based filters \cite{wang2024moderator, kumari2023ablating}. These technique only partially address model biases and irresponsible generation, and may easily  overlook implicit negative concept associations in the model.

More recently, research on diffusion model latent space manipulation and interpretability is gaining prominence in regards to controlling the T2I outputs. Studies such as \cite{haas2024discovering, zhang2024steerdiff}  explore the semantic organization within the latent spaces to understand and intervene in the generation process. Li \etal\cite{li2024self} investigated interpretable directions of target concepts in the latent semantic space. These methods remain effective, however; their underlying discretized treatment of the individual semantic concepts limits their scalability. In this work, we address safe and fair modulation of both embedding and latent spaces without explicitly discretizing the directions, and focusing on their continuous space instead. Our distinctive plug-and-play approach learns an interpretable space for a range of responsible concepts while  conditioning it on the T2I pipeline.  

\begin{figure*}
    \captionsetup[subfigure]{labelformat=nocaption}
    \centering
    \includegraphics[width=\linewidth]{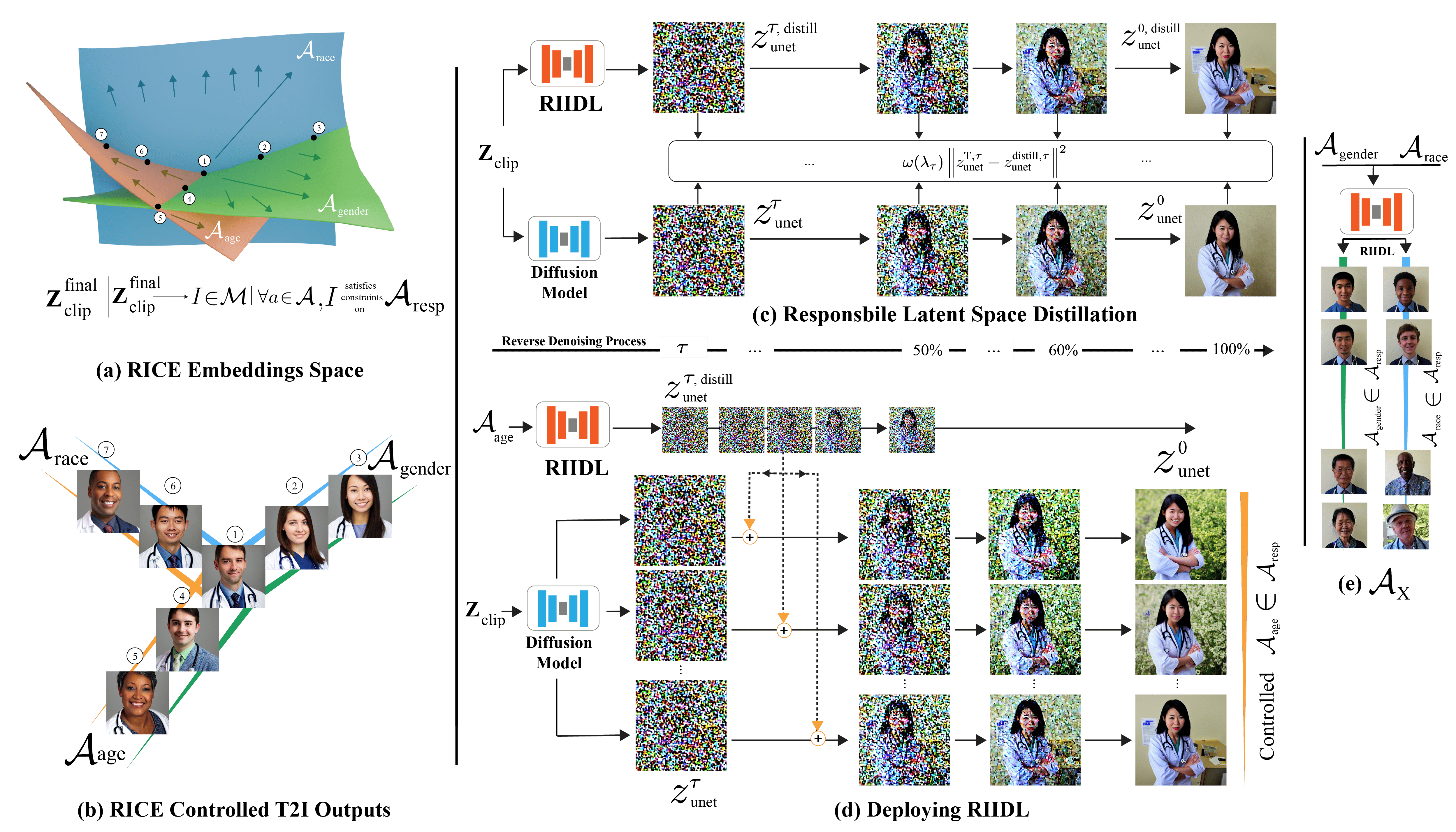}


\begin{subfigure}{0\linewidth}
\caption{}\label{image-a}
\end{subfigure}
\begin{subfigure}{0\linewidth}
\caption{}\label{image-b}
\end{subfigure}
\begin{subfigure}{0\linewidth}
\caption{}\label{image-c}
\end{subfigure}
\begin{subfigure}{0\linewidth}
\caption{}\label{image-d}
\end{subfigure}
\begin{subfigure}{0\linewidth}
\caption{}\label{image-e}
\end{subfigure}
\vspace{-5mm}

\caption{ (a) Illustration of  responsible concept space learned by the RICE module, aiming to generate $z_{\text{clip}}^{\text{final}}$ embeddings along the responsible aspects in $\mathcal {A}_{\text{X}}$. Interconnected fairness related subspaces for race, gender, and age are illustrated. 
(b) Example outputs of RICE controlled T2I process, demonstrating a range of modulation along the concepts, which is possible due to the multi-facet  control enabled by the RICE embedding space. 
(c) Depiction of the distillation process used for the RIIDL module, where the student (RIIDL)  learns from the teacher (Diffusion Model) over $\tau$ time-steps. At each step,  the loss - see Eq.~(\ref{eq:diff}) - aligns the RIIDL module with the aspects of $\mathcal{A}_{X}$.
(d)~At inference, textual embeddings get fed into RIIDL. We illustrate control over $\mathcal{A}_{\text{age}}$. The latent vectors from RIIDL are injected into the T2I Diffusion Model noisy latents. We explore this injection at varying time-step (differentiated vertically in the figure). The best control is observed in the early stages $(\tau = 0~\text{to}~30\%)$. Additional examples provided in the supplementary material. 
(e) Our method allows a scalable control over all concepts covered in $\mathcal A_X$ in a composite manner. 
}

\label{fig:d_diff}
\vspace{-3mm}
\end{figure*}

\vspace{-2mm}
\section{Proposed  Approach}
\label{sec:PA}
\vspace{-1mm}
We present a unique approach for responsible image generation with T2I pipeline that develops add-on modules usable in a plug-and-play manner. The modules are neural models  induced by distilling the target T2I model for a scalable list of responsible sematic concepts addressing fair and safe (non-violent/explicit) image generation. 
We first formally present the problem as perceived in this work.

\vspace{-1mm}
\subsection{Problem Formalization\label{sec:PF}}
\vspace{-1mm}

A text-to-image model, $\Psi (t): t\rightarrow \bar I$, maps a textual prompt $t \in \mathcal T$ to an image $\bar I \in \mathcal M \in \mathbb R^{H\times W \times 3}$, where $\mathcal T$ and $\mathcal M$ are the sets of  possible text prompts and images. Depending on the data used to train the T2I model, the image $\bar I$ may belong to an image subset $\overline{\mathcal R}_{\mathcal A_X} \subset \mathcal M$. Conditioned on ${\mathcal A_X}$,  $\overline{\mathcal R}_{\mathcal A_X}$ denotes the set of irresponsible/inappropriate images as identified by a collection of high-level cultural/social/societal aspects in set ${\mathcal A_X}$. In our  settings, ${\mathcal A_X}$ is a subset of the aspects enlisted in Eq.~(\ref{eq:responsible}). Notice that, $\mathcal A_X \subseteq \mathcal{A}_{\text{resp}}$  implies that $\overline{\mathcal R}_{\mathcal A_X}$ accounts for the fact that $\bar I$ can be inappropriate along more than one aspect - an issue often ignored in the existing literature.
\begin{equation}
\mathcal{A}_{\text{resp}} = \{\mathcal{A}_\text{age}, \mathcal{A}_\text{gender}, \mathcal{A}_\text{race}, \mathcal{A}_\text{safe}\}.
\label{eq:responsible}
\end{equation}
We form $\mathcal{A}_{\text{resp}}$ considering the aspects of responsible image generation accounted in the existing works \cite{li2024self,schramowski2023safe}. Nevertheless, we keep $\mathcal{A}_{\text{resp}}$ extendable  as required. In this work, we define each of the considered aspects using high-level attributes listed below, which also aligns with the existing literature~\cite{li2024self,schramowski2023safe}:    
\( \mathcal{A}_{\text{gender}} = \{\mathbf{a}_{\text{male}}, \mathbf{a}_{\text{female}}\} \), \( \mathcal{A}_{\text{race}} = \{\mathbf{a}_{\text{white}}, \mathbf{a}_{\text{asian}}, \mathbf{a}_{\text{black}}\} \), \( \mathcal{A}_{\text{age}} = \{\mathbf{a}_{\text{young}}, \mathbf{a}_{\text{middle-aged}}, \mathbf{a}_{\text{elderly}}\} \), and \( \mathcal{A}_{\text{safe}} = \{\mathbf{a}_{\text{harassment}}, \mathbf{a}_{\text{sexual}},  \mathbf{a}_{\text{violence}}\} \).

It is worth emphasizing that the attribute sets noted above can also be extended if required. Our approach is not constrained by these sets, except for further training required to account for any additional aspect. Given an $\mathcal A_X$ - as defined by the model user, our objective is to alter the mapping of  the T2I model to  $\Psi_{\text{resp}} (t): t\rightarrow  I$ such that $I \in \mathcal R_{\mathcal A_X} \subseteq \mathcal M$ and $\overline{\mathcal R}_{\mathcal A_X} \cap {\mathcal R}_{\mathcal A_X} = \emptyset$. Here, $\Psi_{\text{resp}}$ is the responsible variant of $\Psi$ with an added functionality that restricts the outputs  of $\Psi$ to $\mathcal A_X$-constrained responsible image space  $\mathcal R_{\mathcal A_X}$. 

\vspace{-2mm}
\subsection{Framework Blueprint}\label{sec:F}
\vspace{-1mm}
Conceptually, a T2I model $\Psi$ is a hierarchical function $\Psi (t) = \mathcal D (\mathcal{E}(t))$, where $\mathcal E(.)$ is a text-encoder - typically CLIP~\cite{kim2022diffusionclip}, and $\mathcal{D}(.)$ is a diffusion model~\cite{rombach2022high}. In this work, we first seek $\Psi_{\text{resp}}$ mentioned in \S~\ref{sec:PF} by devising add-on functionalities for responsible image generation while targeting $\mathcal E(.)$ and $\mathcal D(.)$ individually. In other words, we  aim for  $\mathcal E_{\text{resp}}(.)$ and $\mathcal D_{\text{resp}}(.)$. Both of these potentially `enhanced' sub-models can also be combined  for $\Psi_{\text{resp}}$. We target the text encoder and diffusion model individually because in T2I modeling, both of them are known to effectively encode high-level semantic information in their embedding and latent spaces~\cite{meng2023distillation,haas2024discovering, zhang2024steerdiff}. This provides us the possibility to plug-in our intended add-on modules to any of these models for responsible T2I generation.  

For both of the desired $\mathcal E_{\text{resp}}(.)$ and $\mathcal D_{\text{resp}}(.)$, the  framework underlying our approach remains the same at the conceptual level. We keep the target sub-model ($\mathcal E$ or $\mathcal D$) in the T2I pipeline frozen,  and learn an add-on module (respectively, $\psi_{\mathcal E}$ or $\psi_{\mathcal D}$) by distilling the target sub-model with knowledge distillation \cite{meng2023distillation}. The knowledge is distilled by conditioning the process on the set $\mathcal A_X$ which covers the aspects along which the model needs to be made responsible. Moreover, to ensure that the distilled concepts are disentangled as much as possible, we apply concept whitening~\cite{Chen2020cynthia} to them. The eventual  T2I signal gets modulated  by a simple addition of the signal from the add-on modules for responsible generation. 

We explain the exact procedure for obtaining $\psi_{\mathcal E}$ for $\mathcal E_{\text{resp}}(.)$, and $\psi_{\mathcal D}$ for $\mathcal D_{\text{resp}}(.)$ in \S~\ref{subsec:RICE} and \ref{subsec:RIIDL}, respectively. Combining them for $\Psi_{\text{resp}}$ is explained in \S~\ref{sec:Dual}.






\vspace{-1mm} 
\subsection{Responsible Interpretable Embeddings} \label{subsec:RICE}
\vspace{-1mm}
Our add-on module $\psi_{\mathcal E}$ for $\mathcal E_{\text{resp}}(.)$ is marked by not only \textit{responsible} generation but also \textit{interpretability} of the embedding space distilled from the source $\mathcal E(.)$. Hence, we term the devised module as RICE - Responsible and Interpretable CLIP Embedding. The RICE module treats the original text encoder in the T2I pipeline as a teacher and distills it for the concepts in $\mathcal A_X$ by training $\psi_{\mathcal E}$ while minimizing the Expected value of  the following per-batch loss: 
\begin{equation}
    \mathcal{L}_{\text{KD-clip}} = \frac{1}{|{B}|} \sum_{k=1}^{|{B}|} \left \| z_{\text{clip}, k}^{\text{T}} - z_{\text{clip}, k}^{\text{distill}} \right \|^2,
    \label{eq:KD-CLIP}
\end{equation}
where ${B}$ denotes the batch,  \( z_{\text{clip}, k}^{\text{T}} \) and \( z_{\text{clip}, k}^{\text{distill}} \) are the teacher and distilled student embeddings for the $k^{\text{th}}$ sample, respectively. The $z_{\text{clip}}^{\text{T}}$ embeddings are computed using an automated prompt generation following $\mathcal A_X$. To keep the flow of discussion, we provide  details about that process and the architecture of $\psi_{\mathcal E}$ in the supplementary material. 


At this stage, ${\bf z}_{\text{clip}}^{\text{distll}} \in \mathbb R^{d \times n }$ are the embedding representations conditioned on $\mathcal{A}_X$.
We additoinally apply concept whitening \cite{Chen2020cynthia} to further decorrelate these distilled embeddings using the following transform:
\begin{equation}
    \mathbf{z}_{\text{clip}}^{\text{zca}} = \phi(\mathbf{z}_{\text{clip}}^{\text{distill}})= \mathcal{W}(\mathbf{z}_{\text{clip}}^{\text{distill}}-\mu),
    \label{eq:zca}
\end{equation}
where $ \mu = \frac{1}{n} \sum_{i=1}^{n}z_{i} $ , and $\mathcal{W}_{d\times d}$ is the whitening matrix that obeys $\mathcal{W}^{T}\mathcal{W}=\Sigma^{-1}$. Here, $\Sigma_{d\times d} = \frac{1}{n}(z_{\text{clip}}^{\text{distill}}-\mu {\bf 1}^{\intercal})(z_{\text{clip}}^{\text{distill}} - \mu {\bf 1}^{\intercal})^{\intercal}$ is the covariance matrix. The final embeddings for responsible generation get computed as: 
\begin{equation}
\mathbf{z}_{\text{clip}}^{\text{resp}} = \alpha \cdot \mathbf{z}_{\text{clip}}^{\text{distill}} + (1-\alpha) \cdot \mathbf{z}_{\text{clip}}^{\text{zca}},
    \label{eq:alpha_zca}
\end{equation}
where $\alpha$ is kept for balancing. At inference, the embedding  ${z}_{\text{clip}}$ of the original prompt gets modulated as
\begin{equation}
{z}_{\text{clip}}^{\text{final}} = {z}_{\text{clip}} + \sum_k \gamma_k \mathbf{z}_{k,\text{clip}}^{\text{resp}},
\label{eq:clip_responsible}
\end{equation}
where \( \gamma_k \) provides control over the influence of each concept on the embedding. With this, the embedding of a prompt can be simultaneously modulated to multiple responsible concepts represented in the continuous representation space of $\psi_{\mathcal{E}}$. Fig.~\ref{image-a} illustrates this notion, with Fig.~\ref{image-b} showing the RICE impact on varying the eventual T2I output along different responsible concepts.  

\vspace{-1mm}
\subsection{Responsible Interpretable  Latents}
\label{subsec:RIIDL}
\vspace{-2mm}
In regards to $\mathcal D_{\text{resp}}(.)$, we follow a similar process as adopted in \S~\ref{subsec:RICE} for $\mathcal{E}_{\text{resp}}(.)$, tailoring the former for the diffusion model component $\mathcal D$ of the T2I pipeline. Our add-on module $\psi_{D}$, again a neural model, gets trained using the intermediate latent representations of $\mathcal D$, distilled by conditioning it on $\mathcal A_X$, hence; we  term it RIIDL -  Responsible Interpretable Intermediate Diffusion Latents. 

The latent representation of a diffusion model gets updated for each time-step ($\tau$) of a Markov process denoising the image previously corrupted in the forward diffusion process. For brevity, we refrain from discussing minutiae of diffusion modeling - interested readers are referred to \cite{croitoru2023diffusion}. Here, we focus on the key idea of inducing our plug-in $\psi_{D}$ for the process.
To train $\psi_{D}$, we define the following loss   
\begin{equation}
    \mathcal{L}_{\text{KD-unet}} = \mathbb{E}_{\tau \sim \mathcal{U}[0,1]} \left[ \omega(\lambda_{\tau}) \left\| z_{\text{unet}}^{\text{T}, \tau} - z_{\text{unet}}^{\text{distill}, \tau} \right\|^2 \right],
    \label{eq:diff}
\end{equation}
where $\tau \sim \mathcal{U}[0,1]$ refers to uniform sampling of time-steps, $z_{\text{unet}}^{\text{T}, \tau}$ and $z_{\text{unet}}^{\text{distill}, \tau}$ represent the latent representations of the teacher and student models at time-step $\tau$, $\lambda_\tau$ reflects the SNR of the signal at $\tau$, and $\omega (\lambda_\tau)$ is a pre-specified weighting function. We follow \cite{kingma2021variational} for implementing the SNR and $\omega(.)$. In this setup, $\psi_{D}$ is the student while the T2I  diffusion model is the teacher.





In Eq.~(\ref{eq:diff}), the subscript `unet' emphasizes the U-Net architecture of the student and teacher, which is commonly followed in denoising diffusion models. Details of our student model $\psi_{\mathcal D}$ are provided in the supplementary material. Similar to $z^{\text{T}}_{\text{clip}}$ in Eq.~(\ref{eq:KD-CLIP}),  the $z_{\text{unet}}^{\text T}$ at  $\tau$ is generated based on $\mathcal A_X$, which conditions $\psi_{\mathcal{D}}$ on the responsible concepts being considered by the user.  We further follow the processing of $z_{\text{unet}}^{\text T}$ at a given $\tau$ corresponding to Eq.~(\ref{eq:zca}), (\ref{eq:alpha_zca}) and (\ref{eq:clip_responsible}) to respectively compute $\mathbf{z}_{\text{unet}}^{\text{zca}}$, $\mathbf{z}_{\text{unet}}^{\text{resp}}$, and ${z}_{\text{unet}}^{\text{final}}$ at $\tau$. Fig.~\ref{image-c}  illustrate the distillation process to train $\psi_{\mathcal D}$ for the RIIDL module on receiving the textual embedding input from $\mathcal E$. In Fig.~\ref{image-d}, an illustration is provided for the deployed model, where RIIDL is able to guide the generated image by injecting signals to the early denoising stages of the T2I diffusion model to encourage generation along responsible the concept in $\mathcal A_X$. Figure~\ref{image-e} further shows that extending $\mathcal A_X$ enables simultaneous control over multiple concepts because $\psi_{D}$ gets conditioned on multiple concepts.

\vspace{-1mm}
\subsection{Dual-Space Integration}
\label{sec:Dual}
\vspace{-1mm}
The methods discussed in \S~\ref{subsec:RICE} and \S~\ref{subsec:RIIDL} provide add-on modules $\mathcal{E}_{\text{resp}}(\cdot)$ and 
 $\mathcal{D}_{\text{resp}}(\cdot)$ in the form of plug-and-play models $\psi_{\mathcal E}$ and $\psi_{\mathcal D}$ that can be directly plugged into the text encoder $\mathcal E$ and diffusion model $\mathcal D$ of the T2I pipeline. Whereas both modules are usable standalone, we also integrate them in this work for a comprehensive dual space control on the generative outputs.  
When both modules are integrated into the T2I pipeline, the  overall function of generating a responsible image $I$ from a prompt $t \in \mathcal{T}$ can be expressed as: 
\begin{equation}
   \Psi_{\text{resp}}(t) =  \lambda_{\mathcal{E}} \cdot\mathcal{D} \left( \mathcal{E}_{\text{resp}}(t)\right) + \lambda_{\mathcal{D}} \cdot \mathcal{D}_{\text{resp}}\left(\mathcal{E}(t)\right) , 
   \label{eq:dualspace} 
\end{equation}
where $\lambda_{\mathcal{E}}$ and $\lambda_{\mathcal{D}}$ are weighting factors following the relation $\lambda_{\mathcal{E}}+ \lambda_{\mathcal{D}}=1$.
The integrated  $\Psi_{\text{resp}}(t)$ provides effective control over high-level semantic information in the embedding and latent spaces. It provides a comprehensive plug-and-play method for the T2I pipeline for altering the behavior of $\Psi$ to follow concepts in $\mathcal{A}_X$ for responsible text-to-image generation.

\begin{table*}[t]
\footnotesize 
\setlength\tabcolsep{3pt} 
\centering
\begin{threeparttable}
\caption{Bias generation quantified by deviation ratio ($0 \leq \Delta \leq 1$). Lower values indicate better performance. Results provided  for Gender and Race bias across standard and extended (Gender-Pro/Race-Pro) Winobias datasets~\cite{zhao2018gender}. }
\label{tab:fair_gen_1}
\begin{tabular}{lcccccccccccccccc}

\toprule
\multirow{2}{*}{\textbf{Attribute}} & \multicolumn{4}{c|}{\cellcolor{gray!15}\textbf{Gender}($\downarrow$)} & \multicolumn{4}{c|}{\cellcolor{gray!15}\textbf{Gender-Pro}($\downarrow$)} & \multicolumn{4}{c|}{\cellcolor{gray!15}\textbf{Race}($\downarrow$)} & \multicolumn{4}{c}{\cellcolor{gray!15}\textbf{Race-Pro}($\downarrow$)} \\
\cmidrule(lr){2-5} \cmidrule(lr){6-9} \cmidrule(lr){10-13} \cmidrule(lr){14-17}
& SD~\cite{Rombach2022} & U~\cite{gandikota2024unified} & V~\cite{li2024self} & Ours & SD~\cite{Rombach2022} & U~\cite{gandikota2024unified} & V~\cite{li2024self} & Ours & SD & U~\cite{gandikota2024unified} & V~\cite{li2024self} & Ours & SD~\cite{Rombach2022} & U~\cite{gandikota2024unified} & V~\cite{li2024self} & Ours \\
\midrule
\textbf{CEO}       & 0.92 & 0.28 & 0.15 & \textbf{0.04} & 0.90 & 0.58 & 0.21 & \textbf{0.13}  & 0.38 & 0.13 & 0.22 & \textbf{0.08} & 0.31 & \textbf{0.08} & 0.22 & 0.12  \\
\textbf{Doctor}    & 0.92 & 0.20 & {0.08} & \textbf{0.05} & 0.52 & 0.32 & \textbf{0.10} & \textbf{0.10}  & 0.92 & \textbf{0.07} & 0.26 & \textbf{0.07} & 0.59 & 0.52 & 0.15 & \textbf{0.12}  \\
\textbf{Nurse}     & 1.00 & 0.39 & 0.96 & \textbf{0.04} & 0.98 & 0.84 & 0.43 & \textbf{0.18}  & 0.76 & 0.25 & 0.30 & \textbf{0.07} & 0.39 & 0.79 & \textbf{0.08} & 0.14  \\
\textbf{Receptionist} & 0.84 & 0.38 & 0.88 & \textbf{0.04} & 0.98 & 0.96 & 0.86 & \textbf{0.17}  & 0.88 & 0.10 & 0.36 & \textbf{0.07}& 0.74 & 0.14 & 0.25 & \textbf{0.11}  \\
\textbf{Teacher}   & 0.30 & 0.06 & 0.51 & \textbf{0.04} & 0.48 & 0.16 & \textbf{0.07} & 0.13 & 0.51 & 0.10 & \textbf{0.04} & 0.08 & 0.26 & 0.23 & 0.21 & \textbf{0.10}  \\
\bottomrule
\end{tabular}
\footnotesize
{SD: Standard Diffusion Model, U: Unified Concept Editing \cite{gandikota2024unified}, V: Vector Interpret Diffusion \cite{li2024self}.}
\end{threeparttable}
\vspace{-3mm}
\end{table*}

\begin{figure*}
    \centering
    \includegraphics[width=0.9\linewidth]{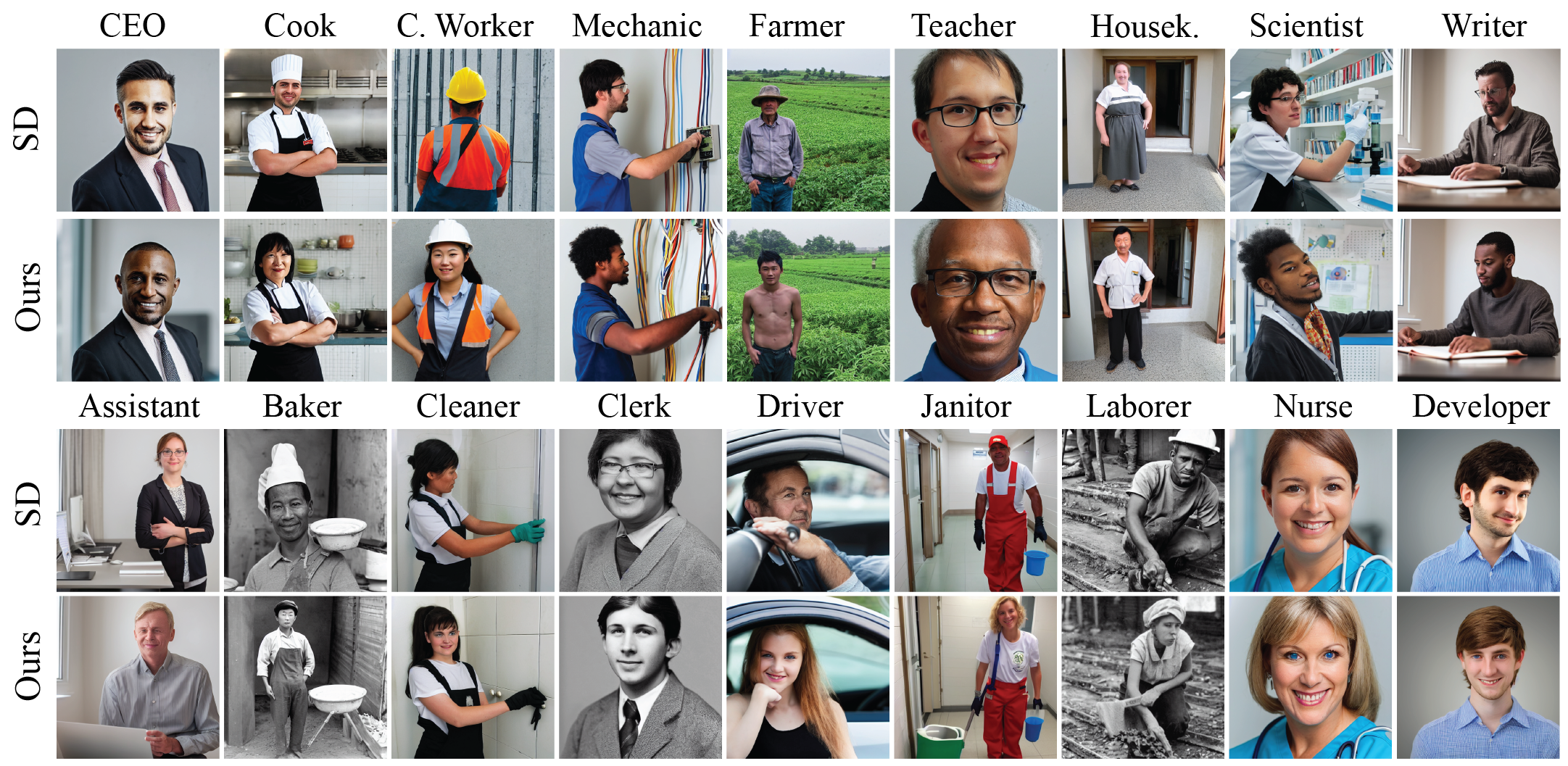}
    \caption{Pair-wise comparison for the Stable Diffusion (SD) baseline and our plugin to it for responsible generation along \textit{age}, \textit{gender} and \textit{race}. SD images contain stereotypical associations to the profession, which get removed by our method by often accounting for more than one responsible attributes, while maintaining high image quality.}
    \label{fig:compare_sd}
    \vspace{-3mm}
\end{figure*}

\begin{table*}[t]
    \centering
    \footnotesize 
    \setlength{\tabcolsep}{3pt} 
    \vspace{-2mm}
    \caption{Debaising performance across 5 randomly selected professions and the average of all 35 professions in Winobias dataset \cite{zhao2018gender}. The metric $\Delta = 0$ indicate ideal debaising. Our method shows the lowest average deviation compared to previous approaches.}
    \vspace{-3mm}
    \begin{tabular}{lcccccc}
        \toprule
        \cellcolor{gray!15}\textbf{Profession} & \cellcolor{gray!15}\textbf{Original-SD} & \cellcolor{gray!15}\textbf{Concept Algebra}~\cite{wang2023concept} &\cellcolor{gray!15} \textbf{Debias-VL}~\cite{chuang2023debiasing} & \cellcolor{gray!15}\textbf{TIME}~\cite{orgad2023editing} & \cellcolor{gray!15}\textbf{TIMEP}~\cite{gandikota2024unified} & \cellcolor{gray!15}\textbf{Ours} \\ 
        \midrule
        Librarian & 0.86 {\scriptsize$\pm$ 0.06} & 0.66 {\scriptsize$\pm$ 0.07} & 0.34 {\scriptsize$\pm$ 0.06} & 0.26 {\scriptsize$\pm$ 0.05} & 0.35 {\scriptsize$\pm$ 0.01} & \textbf{0.04 {\scriptsize$\pm$ 0.02}} \\ 
        Teacher   & 0.42 {\scriptsize$\pm$ 0.01} & 0.46 {\scriptsize$\pm$ 0.00} & 0.11 {\scriptsize$\pm$ 0.05} & 0.34 {\scriptsize$\pm$ 0.06} & 0.07 {\scriptsize$\pm$ 0.06} & \textbf{0.05 {\scriptsize$\pm$ 0.01}} \\ 
        Sheriff   & 0.99 {\scriptsize$\pm$ 0.01} & 0.38 {\scriptsize$\pm$ 0.22} & 0.82 {\scriptsize$\pm$ 0.08} & 0.22 {\scriptsize$\pm$ 0.05} & 0.10 {\scriptsize$\pm$ 0.05} & \textbf{0.06 {\scriptsize$\pm$ 0.03}} \\ 
        Analyst   & 0.58 {\scriptsize$\pm$ 0.12} & 0.24 {\scriptsize$\pm$ 0.18} & 0.71 {\scriptsize$\pm$ 0.02} & 0.52 {\scriptsize$\pm$ 0.03} & 0.13 {\scriptsize$\pm$ 0.05} & \textbf{0.07 {\scriptsize$\pm$ 0.02}} \\ 
        Doctor    & 0.78 {\scriptsize$\pm$ 0.04} & 0.40 {\scriptsize$\pm$ 0.02} & 0.50 {\scriptsize$\pm$ 0.04} & 0.58 {\scriptsize$\pm$ 0.03} & 0.41 {\scriptsize$\pm$ 0.08} & \textbf{0.06 {\scriptsize$\pm$ 0.01}} \\ 
        \midrule
        Average & 0.67 {\scriptsize$\pm$ 0.01} & 0.43 {\scriptsize$\pm$ 0.01} & 0.55 {\scriptsize$\pm$ 0.01} & 0.44 {\scriptsize$\pm$ 0.00} & 0.31 {\scriptsize$\pm$ 0.00} & \textbf{0.05 {\scriptsize$\pm$ 0.02}} \\ 
        \bottomrule
    \end{tabular}
    \vspace{-3mm}\label{tab:debiasing_performance}
\end{table*}


\begin{table}[t]
    \centering
    \footnotesize
     \setlength{\tabcolsep}{3pt} 
     \caption{Proportion of images classified as inappropriate on I2P benchmark \cite{schramowski2023safe}. Lower values are more desirable.}    \renewcommand{\arraystretch}{1} 
     \vspace{-3mm}
    \begin{tabular}{lccccc}
        \toprule
        \cellcolor{gray!15}\textbf{Category} & \cellcolor{gray!15} \textbf{SD} &  \cellcolor{gray!15} \textbf{V-SD~\cite{li2024self}} & \cellcolor{gray!15}  \textbf{SLD~\cite{schramowski2023safe}} & \cellcolor{gray!15}  \textbf{ESD~\cite{gandikota2023erasing}} & \cellcolor{gray!15}  \textbf{Ours} \\
        \midrule
        Sexual       & 0.39 & 0.23 & 0.16 & 0.19 & \textbf{0.15} \\
        Violence     & 0.45 & 0.31 & \textbf{0.22} & 0.42 & 0.24 \\
        Hate         & 0.42 & 0.30 & 0.19 & 0.34 & \textbf{0.18} \\
        Harassment   & 0.35 & 0.19 & \textbf{0.16} & 0.28 & 0.19 \\
        Illegal      & 0.35 & 0.24 & 0.18 & 0.34 & \textbf{0.14} \\
        Shocking     & 0.53 & 0.38 & \textbf{0.27} & 0.43 & 0.28 \\
        Self-harm    & 0.45 & 0.29 & 0.20 & 0.36 & \textbf{0.20} \\
        \midrule
        Avg. & 0.41 & 0.27 & \textbf{0.20} & 0.32 & \textbf{0.20} \\
        \bottomrule
    \end{tabular}
    \label{tab:i2p_benchmark}
    \vspace{-5mm}
\end{table}

\vspace{-2mm}
\section{Evaluation}
\label{sec:evaluation}
\vspace{-2mm}
Our method is evaluated for a variety of concepts related to fair and safe image generation. 
The used evaluation protocols are based on standardized benchmarks. 
For reproducibility, further details on  implementation, dataset descriptions, and configurations of hardware and software libraries are also provided in the supplementary material. 
Our experiments use the Stable Diffusion v1.4 model~\cite{Rombach2022}.
We compare with SLD \cite{schramowski2023safe}, ESD \cite{gandikota2023erasing}, UCE \cite{gandikota2024unified}, and Vector-SD \cite{li2024self}, using  WinoBias \cite{zhao2018gender} and I2P \cite{schramowski2023safe} datasets, which contain  fairness-sensitive and safety-critical scenarios. 


\noindent{\bf Fair Generation Evaluation:}
In regards to the fair generation, we evaluate  performance across professions with documented gender and racial biases. The  dataset for this evaluation is WinoBias \cite{zhao2018gender}, which comprises 35 professions historically associated with gender bias. Fairness is quantitatively assessed using the deviation ratio \(\Delta\) as in literature \cite{orgad2023editing, chuang2023debiasing, li2024self}, which measures the disparity in generated representations across specified responsible aspects by comparing the model’s output distribution against an expected baseline. Formally, \(\Delta\) is defined as: $\Delta = \frac{\max_{a \in \mathcal{A}} \left| \frac{N_a}{N} - \frac{1}{|\mathcal{A}|} \right|}{1 - \frac{1}{|\mathcal{A}|}}$, where \(\mathcal{A}\) is the set of attributes within a responsible aspect (e.g., gender or race), \(N\) represents the total number of generated images, and \(N_a\) denotes the count of images for which the highest probability attribute prediction corresponds to attribute \(a\).
For extensive evaluation, we further employ “challenging prompts” \cite{li2024self} in our experiments, which are known to produce bias representations towards male depictions \cite{gandikota2024unified}. 

\vspace{0.5mm}
\noindent{\bf Safe Generation Evaluation:}
We also evaluate the generated content against the standard criteria  for identifying harmful or culturally sensitive outputs \cite{schramowski2023safe,li2024self}. Images are flagged as inappropriate if they depict or imply content associated with  \textit{sexual content, self-harm, hate, illegal activity, shock, harassment, or violence}. Our evaluation  integrates both the NudeNet detector\footnote{https://github.com/notAI-tech/NudeNet} and Q16 classifier \cite{schramowski2022can}, as per common practices in responsible generation assessment, to systematically flag content deemed inappropriate~\cite{schramowski2023safe,li2024self}. We examine a diverse range of images to rigorously assess robustness across varied outputs. 

\vspace{-2mm}
\section{Experimental Results}
\label{sec:dual_space}
\vspace{-1.5mm}
\noindent{\bf Fair Generation:} In Table~\ref{tab:fair_gen_1}, we present the bias quantification results using the deviation ratio (\(0 \leq \Delta \leq 1\)) across gender and racial attributes, evaluated under both the standard (Gender and Race) and extended (Gender-Pro and Race-Pro) settings of the WinoBias dataset~\cite{zhao2018gender}. Lower values of \(\Delta\) indicate reduced demographic bias.
Overall, our method consistently achieves highly competitive results. 
Even under the Gender-Pro setting, where prompts are specifically crafted to amplify stereotypical associations, our method maintains its robustness.  
For racial bias, our method similarly maintains its performance. 

Table~\ref{tab:debiasing_performance} further quantifies the debiasing efficacy of the our approach across professions with known demographic skew, displaying results for five professions and the average deviation across all 35 professions in the WinoBias~\cite{zhao2018gender}. Results show a substantial bias reduction by our method. 
Consistent bias reduction demonstrates that bias mitigation is achieved by our method independent of profession-specific attributes. Averaged over all professions (last row), our method provides a significant gain over the baseline methods. 
Such efficacy of our method comes from its ability to enable debiasing using a collective set of attributes in $\mathcal A_X$ and modulate the T2I pipeline to generate images responsibly considering all those attributes.

To show how this manipulation varies the images generated by Stable Diffusion (SD), in Fig.~\ref{fig:compare_sd} we provide representative examples. The figure shows a range of professions for which pairs of images are generated using the same seeds. The SD images show stereotypical associations to the professions. Our method is plugged in to SD with set $\mathcal{A}_X$ collectively containing \textit{race}, \textit{gender} and \textit{age} attributes. It is observable in the provided representative examples that images get modulated towards these responsible concepts, often incorporating more than one responsible concept. We also provide further results in the supplementary material.

\begin{figure*}[!ht]
    \centering
    \includegraphics[width=0.9\linewidth]{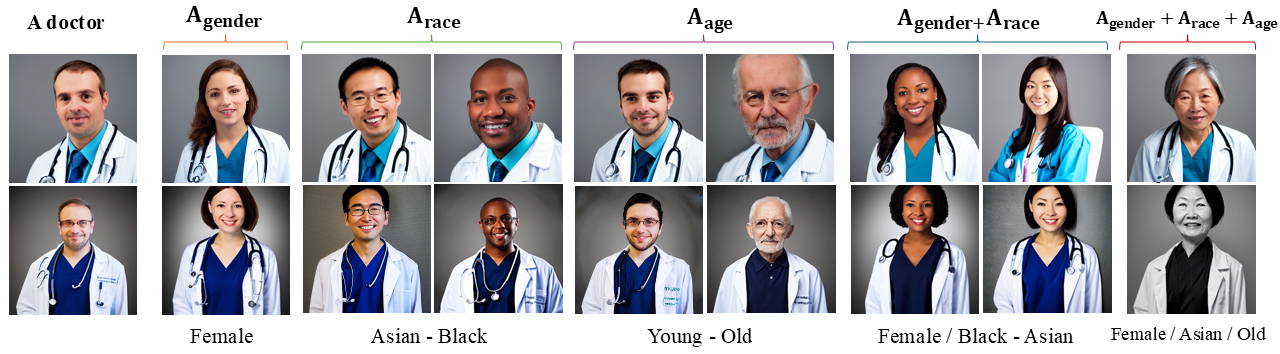}
    \vspace{-3mm}
    \caption{Controlled composite responsible generation using the proposed method. By using different concepts in \(\mathcal{A}_X\)  in Eq.~(\ref{eq:responsible}), and employing dual space control in (\ref{eq:dualspace}), our technique can enable responsibility along single or multiple concepts, as desired. 
    Provided the prompt “A doctor” to Stable Diffusion, alignment is achieved for a target concept composition (top label) for the attributes noted at the bottom of each image set. See supplementary material for more examples.}
    \label{fig:multi}
    \vspace{-4mm}
\end{figure*}

\noindent\textbf{Safe Generation:} To benchmark safe and appropriate nature of the generated content, we evaluate our method on the I2P~\cite{schramowski2023safe}, which classifies generated images across categories  covering a range of inappropriate content. 
Table \ref{tab:i2p_benchmark} benchmarks   our method against the state-of-the-art methods following standard protocol~\cite{schramowski2023safe}, \cite{gandikota2023erasing}.     
As seen, our technique significantly reduces  inappropriateness of the content  across all categories. Generally, maintaining superior performance against the methods. On average, our performance is similar to SLD~\cite{schramowski2023safe}, which is a dedicated `safe generation' method. 

\begin{figure}[!ht]
    \centering
    \includegraphics[width=\linewidth]{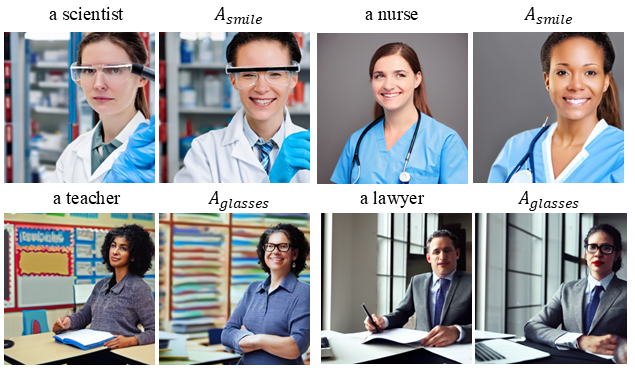}
    \vspace{-3mm}
    \caption{Examples of extending $\mathcal A_{\text{resp}}$ with additional attributes - possibly unrelated to core responsible concepts. Extensions with \textit{smile} and \textit{glasses} are shown. In addition to varying images along responsible concepts of \textit{age}, \textit{race} and \textit{gender}, our method is able to  seamlessly incorporate the  additional concepts in images.}
    \label{fig:scale_concept}
    \vspace{-5mm}
\end{figure}

\vspace{-2mm}
\section{ Further Results \& Discussion}\label{subsec:ablation}
\vspace{-1mm}


\noindent \textbf{Multi-Concept Composition and Control:} Our technique allows responsible generation with multi-concept composition. Multiple concepts can compose $\mathcal{A}_X$, which enables the hierarchical function \(\Psi_{\text{resp}}\)  combine them in latent and embedding space of the T2I pipeline. 
Figure~\ref{fig:multi} demonstrates this capability within \(\Psi_{\text{resp}}\), showing how distinct responsible aspects, such as age, gender, and race; can be  independently learned and then composed to produce varied yet coherent outputs. By manipulating combinations within \(\mathcal{A}_X\), we demonstrate an excellent  control achieved with interpretable variations across the composite attribute space, effectively balancing diversity with responsible aspects. 

\noindent \textbf{Scalability of Concept Spaces:} In our original experiments, the responsible concept set \(\mathcal{A}_{\text{resp}}\) in our framework, as defined in Eq.~(\ref{eq:responsible}), included core attributes of $\mathcal{A}_{\text{race}}, \mathcal{A}_{\text{gender}}, \mathcal{A}_{\text{age}}$, and $\mathcal{A}_{\text{safe}}$. This was only to benchmark the method with the existing approaches. 
Our design allows the ``responsible" set to be extendable.  This scalability allows for seamlessly handling additional nuanced attributes in images, such as facial expressions (e.g., smiling) and accessories (e.g., glasses). Such concepts can act as  supplementary attributes with the existing attributes, enabling complex yet interpretable multi-aspect control. In Fig.~\ref{fig:scale_concept}, we provide representative examples of scaling up the considered concepts set. It leads to  combining the responsible subspaces of the core responsible features with subspaces for smiling and glasses. It is observable that our framework is scalable to not only core responsibility related concepts, but to any semantically meaningful concepts. We also provide additional results in the supplementary material.

\noindent \textbf{Individual Modules:} 
Our method takes advantage of two modules, namely; of RICE (\S~\ref{subsec:RICE}) and RIIDL (\S~\ref{subsec:RIIDL}).
To assess the specific contribution of each component within the responsible generation framework, we perform a systematic ablation by individually removing and adjusting key modules. This analysis allows selectively varying the influence both modules on the output. 
When only RICE operates, modifications originally occur in the embedding concept space for the T2I pipeline, facilitating transitions from base to responsible concept spaces. 
The RIIDL module contribute an additional layer of fine-grained control, allowing nuanced and context-sensitive adjustments within the intermediate latents during diffusion. This capability supports precise tuning of details without overriding the global context, complementing the broader adjustments. In tandem, these modules establish a dynamic balance within the dual-space intervention strategy, where RICE provides pre-diffusion global control, and RIIDL governs the responsible aspects during the denoising process. 
Whereas we provide further results supporting these point in the supplementary material, in
Fig.~\ref{eq:diff} a curtailed representative example is provided to show the visual effects. 

\begin{figure}
    \centering
    \includegraphics[width=\linewidth]{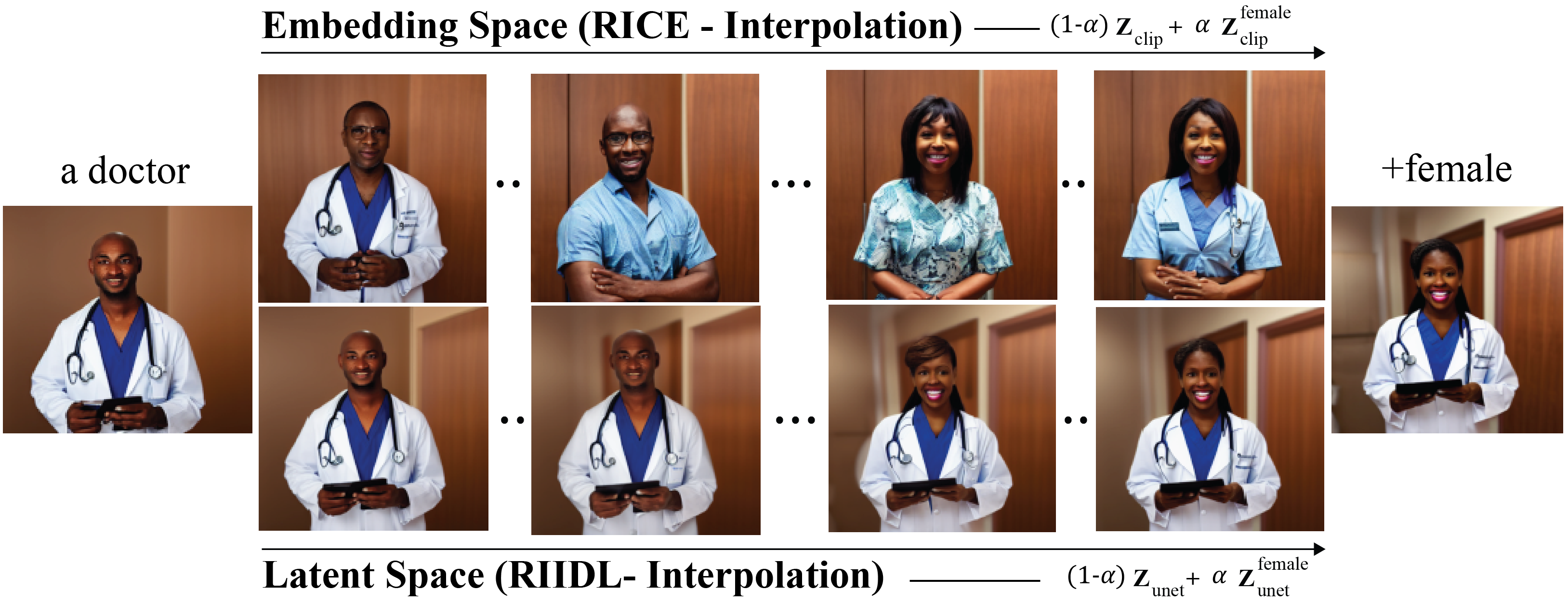}
    \vspace{-4mm}
    \caption{Representative curtailed example of transitions caused by RICE and RIIDL to an output. An image of `a doctor' is generated and `female' concept is added individually by the modules. Increasingly changing the influence of the added concept representation provided by each module leads to unique alteration to the generated output. }
    \label{fig:interpolation}
    \vspace{-5mm}
\end{figure}

In the figure, we generate images using prompt `a doctor' and with an added concept `female' from the individual modules. The `female' concept  representation ${\bf z}^{\text{female}}$ provided by the modules is added to the original corresponding representations with an increasing degree of influence on the output. 
It can be observed that for RICE, this results in more prominent shifts in appearance and attire, indicating a significant control. For RIIDL, the latent space provides more refined transitions with subtle and smoother progression of the visual features. On one hand, these results demonstrate the complementary nature of the proposed modules, on the other; they establish the effectiveness of the modules as stand-alone add-ons. 

\vspace{-2mm}
\section{Conclusion}
\vspace{-1mm}
This research presented a unique approach to control generative outputs of text-to-image (T2I) models in `fair and safe' domain. Our method induces two plug-and-play modules that can be plugged into the textual embedding space and diffusion latent space of the T2I pipeline. The modules are neural models learned by distilling the text encoder and diffusion model of the original pipeline while conditioning the knowledge distillation on a pre-defined responsible concepts set, as desired by the user. Our modules also account for semantic interpretability by leveraging concept whitening. We demonstrate that both modules can explicitly contribute to responsible image generation in a complementary manner. Our extensive benchmarking with state-of-the-art methods for fair and safe image generation on two standard datasets show highly promising results. We also show that our method can conveniently deal with multiple concepts in the responsible image space while being seamlessly extendable to more broader concepts.

\vspace{0.5mm}
\noindent{\bf Limitations and ethics concerns:} A potential shortcoming of our method is that, to enable a comprehensive control over the generative output, it employs multiple hyper-parameters, e.g., $\lambda$'s in Eq.~(\ref{eq:dualspace}). Tuning these hyper-parameters can potentially be seen as a limitation of our approach. However, the main purpose of allowing  these adjustable hyper-parameters in our design is to provide a finer control to the user. It is convenient to adjust their values because the controls they provide have clear high-level interpretations.  We provide guidelines on the hyper-parameter value adjustment in the supplementary material. From the ethics viewpoint, although our aim is to ensure responsible outputs; a potential misuse of exploiting the strong output control of our technique can be by replacing the $\mathcal A_{\text{resp}}$ with another set $\mathcal A_{\text{irresp}}$. In $\mathcal A_{\text{irresp}}$, unsafe and unfair concepts can be included. This can have an opposite impact of our technique. Hence, our method should only be used when the user has complete control over the set $\mathcal A_{\text{resp}}$. 


\noindent{\bf Acknowledgment:} Naveed Akhtar is a recipient of the Australian Research Council Discovery Early Career Researcher Award (project \# DE230101058) funded by the Australian Government. This work is also partially supported by Google Research Scholar Program Award.

{
    \small
    \bibliographystyle{ieeenat_fullname}
    \bibliography{main}
}









\end{document}